# Embedding Data within Knowledge Spaces

James D. Myers, Joe Futrelle, Jeff Gaynor, Joel Plutchak, Peter Bajcsy, Jason Kastner, Kailash Kotwani, Jong Sung Lee, Luigi Marini, Rob Kooper, Robert E. McGrath, Terry McLaren, Alejandro Rodriguez, Yong Liu

National Center for Supercomputing Applications, University of Illinois at Urbana Champaign

**Abstract**

The promise of e-Science will only be realized when data is discoverable, accessible, and comprehensible within distributed teams, across disciplines, and over the long-term – without reliance on out-of-band (non-digital) means. We have developed the open-source Tupelo semantic content management framework and are employing it to manage a wide range of e-Science entities (including data, documents, workflows, people, and projects) and a broad range of metadata (including provenance, social networks, geospatial relationships, temporal relations, and domain descriptions). Tupelo couples the use of global identifiers and resource description framework (RDF) statements with an aggregatable content repository model to provide a unified space for securely managing distributed heterogeneous content and relationships. The Tupelo framework includes an HTTP-based data/metadata management protocol, application programming interfaces, and user interface widgets which have been incorporated into NCSA's portal and workflow tools and is a key component in recent work creating dynamic digital observatories (digital watersheds) that combine observational and modeled information. Tupelo also supports specialized indexes and inference logic (computation) relevant to metadata including geospatial location and provenance. This additional capability creates a powerful knowledge space that can map between disciplinary conceptual models and between the storage and data organization choices made by different e-Science organizations.

## Introduction

E-Science and Cyberinfrastructure are often described in terms of new resources and capabilities that will be accessible by researchers (NSF, 2003). However, the vision that such capabilities will enable new research and promote faster transfer of results into practical application assumes that it will be possible for researchers to manage more complex computation, deeper analysis and synthesis of more data, and more interaction with colleagues, i.e. that knowledge transfer will become faster and knowledge management more automated.

Today, scientific data is often managed in relatively static collections with minimal contextual metadata, making it difficult for scientists to understand how to use it. Analytic and computational scientific processes are managed largely in an *ad hoc* manner (e.g., with scripting languages) or using applications and workflow tools, which typically either do not record the details of data processing (data provenance) or do so in internal stores, rather than in the data repositories that serve as the source and/or destination for the results of those processes, with the result that data provenance is not available. The disconnect between processes and data means that creating an automated e-Science environment, capable of reproducing experiments and allowing evolution of analytical processing, requires custom programming or complex manual processes in which the scientist must work with heterogeneous tools with little integration. At the same time, notes and discussions that take place during a scientific project are managed in e-mail or collaboration systems that are typically also dis-integrated from the scientific work itself, so that scientists looking for the collaborative context of a particular project activity typically have to use separate tools to recover notes and messages on the one hand and workflows and datasets on the other. By the time results of a project or experiment are published, most traces of the original process and data are inaccessible to the reader, and only the paper's narrative—highly compressed and unsuitable for machine consumption—provides any information about them, making it difficult to go from paper back to a working research capability (James D. Myers, Chappell, Elder, Geist, & Schwidder, 2003). These barriers to the seamless integration of data and process beyond the scale of a single tool or database limit the utility of current e-Science approaches. Removing them will be a key challenge in community-scale projects such as the environmental observatories now being pursued with the U.S. National Science Foundation (Robertson, 2008). Central to removing these barriers will be semantic web technologies (De Roure & Hendler, 2004), which are currently being used in an e-Science context by a wide range of projects (see references at www.semanticgrid.org), including efforts that maintain data and process connections from laboratory to reference database (Frey et al., 2006). However, the remaining gap between the requirements of e-Science and existing semantic repositories and tools has led us to develop the approach described in this paper. We have begun to develop tools and approaches that embed scientific data in distributed "knowledge spaces". Knowledge spaces provide a single uniform mechanism for accessing data, rich contextual metadata, and inferred or computed information. They use explicit semantic metadata representations of scientific data and processes and enable tracking data and process evolution, the association of heterogeneous artifacts and processes (e.g., notes, literature, experimental apparatus) with data, and the virtual organization (VO) scale implementation of shared semantic contexts such as spatiotemporal coverage and domain-specific ontologies, across a distributed securable context. Knowledge spaces agument semantic web technologies such as RDF (W3C, 1999) and OWL (W3C, 2003) with techniques from Grid Computing (Foster & Kesselman, 1998), scientific workflow systems, and digital libraries to provide "semantic content management" and serve as an integration mechanism across desktop and web-based tools and between personal, group, and public subspaces.

Knowledge spaces extend the idea of content management as represented by standards such as the Java Content Repository (JCR) API (JSR 170 Expert Group, 2005) and WebDAV (Goland, Whitehead, Faizi, Carter, & Jensen, 1999) to incorporate explicit semantics. The value of WebDAV and content management for e-Science has been widely recognized (De Roure et. al., 2001) and is apparent in our own efforts (CMCS, 2007), (J. D. Myers, Spencer Jr, & Navarro, 2006). In this paper, we present NCSA's Tupelo middleware, which implements a semantic content abstraction, and discuss how its unique blend of semantic web and content management functionality within a highly extensible architecture enables interacting VO-scale knowledge spaces.

## Background

Most digital scientific data is part of the "deep web", managed in databases and collections that are neither widely accessible nor organized in a way that can be apprehended without domain-specific or even collection-specific code. To address this problem we have worked to apply best-practice data management strategies to scientific data, including digital library technologies (McGrath, Futrelle, Plante, & Guillaume, 1999), content management systems (James D. Myers et al., 2004), and institutional repositories (Habing, Pearce-Moses, & Surface, 2006). Because most of these technologies primarily focus on managing static data collections, we have developed techniques to integrate them into scientific work processes, including data acquisition systems (NEES, 2003) and sensor networks (Liu et al., 2006), scientific notebooks and collaboration (James D. Myers et al., 2004), and digital curation and preservation (Dubin, Plutchak, & Futrelle, 2006). These complex integration problems have a parallel in developments outside of e-Science such as syndication, mashups (Feiler, 2008), and reflective middleware (Myers and McGrath, 2007), all of which enable content-driven applications and websites to be deployed and integrated with existing tools. Common across these threads is the recognition that e-Science, like enterprise and VO-scale endeavors in industry, has such a wide variety of content types and tools that manually integrating each tool with each data type is impractical.

Our experience has shown that supporting VOs requires "semantic content management" that blends traditional CMS capabilities with the semantic web, so that distributed tools can reliably interpret distributed, heterogeneous data according to explicit domain semantics and automated conversions versus the use of stove-pipe data stores and specialized format conversion code. Best-practice content management tools and APIs such as institutional repositories (IR) (e.g., Fedora (fedora-commons.org, 2008), DSPACE (dspace.org, 2008) ) and CMSs (e.g., Jackrabbit (Apache, 2006), Drupal (drupal.org, 2008)) do not meet this requirement because they typically serve only to make each "island" of data easier to access. Where they do provide some means of integrating multiple collections (e.g., OAI-ORE (Lagoze et al., 2007), institutional repositories) it is typically limited to simple aggregation or syndication of "archival information packages" (AIP) (Lavoie, 2004). Because AIPs and similar structures enforce a single level of granularity on any given data collection and organize information into a closed structure, IR's and CMS's cannot adequately represent the semantic equivalence between two sets of packages that represent the same information organized differently, making complex interrelationships such as social networks and process abstractions awkward to represent without specialized code. Furthermore, in an IR or CMS the contextual metadata that would be required for an application or user to make sense of the heterogeneous content of a scientific data collection (e.g., notes, datasets, literature, code) is often accessible only in its native format or structure, or in a "dumbed-down" (DCMI Usage Board, 2007) form based only on commonly-used attributes and properties.

Techniques exist to address the problem of integrating heterogeneous metadata, most notably semantic web technologies (RDF and OWL), but these are often difficult to deploy in content management systems or institutional repositories because the structural assumptions made by those technologies (e.g., data is organized into a single hierarchy, a property can only have one value, each entity has the same set of properties) are often incompatible with the semantic web's non-hierarchical, open model. For example, a WebDAV (Goland et al., 1999) server assumes that an entity such as a document can only be addressed, accessed, and modified using that WebDAV service, in effect making the service the "owner" of that entity. This is a poor fit for e-Science, where, for example, multiple agents in a sensor network may process measurements before the data are recorded in a central database and where multiple parties may continue analyzing the data after exporting it to their local systems. The tight link between storage location, identifier, and access mechanism makes it extremely difficult to gather information generated by independent parties into a single description.

While RDF and OWL provide a representational framework for distributed, heterogeneous metadata descriptions, they do not prescribe means of managing and accessing collections of RDF descriptions (although a query language, SPARQL, has recently been specified (Prud'hommeaux & Seaborne, 2006)). As a result, a diverse universe of RDF tools and technologies have been developed (e.g., Jena (JENA, 2003), Sesame (openrdf.org, 2008), Mulgara (mulgara.org, 2007)). Although RDF data is portable across virtually all of these tools, the tools themselves are not integrated except in limited ways; moving RDF data from one RDF triple store such as Sesame into another one, such as Jena, requires writing code against both API's. Some proposals have been made to address these problems for service-oriented architectures (SOAs), but implementations are scarce. These include SPARQL-DAP (for querying only) (Clark, Feigenbaum, & Torres, 2008) and URIQA (for both writing and querying) (Stickler, 2004). URIQA is notable in that, analogous to WebDAV and unlike most semantic web tools, it combines support for data and metadata within a single component.

## Tupelo Semantic Content Management Middleware

The Tupelo semantic content management system was originally developed for the NEESgrid earthquake science collaboratory (NEES, 2003) and further developed as part of the Open Grid Computing Environment project (Alameda et al., 2007) and NCSA's Digital Synthesis Framework (TRECC 2008). Tupelo blends ideas from content management systems, grid computing, and the semantic web to provide desktop-to-grid access to semantic metadata and data resources. It defines a WebDAV-style protocol for managing data and metadata and defines an aggregatable context mechanism that allows composition – access through one context to multiple underlying contexts. It provides a low-level client-side API as well as an API for object-oriented interaction with content and a number of Java and web interface components for displaying content in tables, trees, and graphs. On the back end, Tupelo provides a middleware library implementing the access protocol, context mechanism, access control, and related functionality.

Tupelo's context mechanism allows heterogeneous context implementations to be arranged to provide aggregation, mirroring, and failover across multiple Tupelo repositories that appear to the caller as a single data and metadata resource. Context implementations are provided for several leading semantic web databases, such as Sesame and Jena, as well as widely available storage mechanisms and protocols including file systems, databases, SSH, WebDAV, and RSS. Tupelo's context mechanism can also be used to support a plug-in capability analogous to WebDAV's "managed properties" to support the additional indexing, inference, and computation required to provide server-side management of specific types of metadata. Such a mechanism can be used to support operations such as transitive closure (McGrath and Futrelle, 2007) required by the Open Provenance Model (OPM) (Moreau et al., 2007) as well as to compute derived relationships such as the geospatial relationship of being "in" a region derived from latitude and longitude or to implement a streaming data abstraction over discrete storage. In addition to Tupelo, we have also developed a number of Tupelo-aware tools including a portal and a workflow engine. Our ability to easily link these tools into broader frameworks supporting social networking and data and process publication, demonstrate the potential of a semantic content management abstraction in supporting e-Science.

## Architecture

**Contexts**

Tupelo's internal architecture was informed by the Java COG Kit (Laszewski, Foster, Gawor, & Lane, 2001), which provides a generic task abstraction enabling applications to perform grid executions and data transfers on heterogeneous underlying services. In Tupelo, data and computational resources are encapsulated in Contexts, which are responsible for performing Operators such as writing data or performing a query. Tupelo-aware applications access data, metadata, and services through Contexts. In turn Contexts negotiate with each other and with underlying services, processes, or storage resources to perform operations. Two classes of operations are provided as the "kernel" of Tupelo's functionality: metadata operations, including asserting and retracting RDF statements and searching for RDF statements that match simple queries; and data operations, including reading, writing, and deleting binary large objects (BLOB's), each of which is identified with a URI reference.

Other operations may be defined to extend Tupelo's capabilities. Limiting the set of core operations as we have done eases interoperability between Context implementations and simplifies integration of new data stores while still providing a sufficient basis for building more complex operations. Unlike SOA messages, Tupelo operations are stateful objects that are modified as a result of being performed, so that chains or hierarchies of Contexts can transform them before and after they are performed in order to provide additional capabilities such as validation, logging, notification, and the application of rules. Since operators can be defined in terms of one another, Contexts that only support simple operations can be made to perform more complex operations through the use of a wrapper Context that decomposes complex operations into simpler ones. For example a simple Context that can only parse an RDF/XML file can be made to perform queries by decomposing the query operation into an iteration operation over the RDF statements in the file, a write operation to a more capable delegate Context, and a query operation on the delegate.

A variety of Context implementations are provided which demonstrate the application of this declarative approach to interacting with semantic content. Some Contexts wrap existing storage and retrieval mechanisms such as filesystems, databases, and RDF triple stores. Others translate live data (e.g., from an RSS feed) into RDF whenever a query or metadata read operation is performed. Others act as delegates to sets of child Contexts for the purposes of mirroring and failover or combining results from operations performed against many children. And others provide Tupelo operations on widely-used client/server protocols such as HTTP and WebDAV.

Tupelo can combine heterogeneous and otherwise uncoordinated resources into a single Context, which can be used to integrate data, metadata, and provenance from multiple providers into a single coherent representation. For example, a Context can associate workflows and data maintained locally with a paper served from a central server. A Context can also support local annotation and tagging of non-local information, i.e. data stored in a Tupelo-wrapped relational database that does not support annotation directly, and, similarly, the maintenance of a local notebook referencing remote group and reference data. These examples represent two general cases – where multiple Contexts store different metadata about the same resources and where Contexts store different related resources. Further, because Contexts can be created on both the server and via the client library, choices for aggregation can be made by VOs and/or by individual applications. Applications could, for example, be configured to use a local Tupelo repository with data mirrored to the VO server.

**Protocol**

Tupelo also provides an HTTP client/server protocol that is compatible with Nokia's URIQA protocol as well as providing additional query capabilities. In analogy with the base WebDAV protocol, which adds actions for getting and setting metadata (PROPFIND, PROPPATCH) to the standard GET and PUT of HTTP, URIQA and Tupelo add get, set, and delete operations (MPUT, MGET, MDELETE), (NCSA 2008). More broadly, while URIQA and the Tupelo Server protocol use WebDAV's approach of extending HTTP, they combine it with the global addressing and explicit semantic capabilities of RDF, allowing clients to represent not just data objects and their attributes but also concepts and ontologies whose identifiers were minted elsewhere. For this reason, Tupelo allows metadata operations to include the subject of a triple (versus WebDAV's assumption that all key/value pairs apply to the resource identified by the URL used in the PROPPATCH operation). Tupelo and URIQA also relax WebDAV's requirements that objects be organized into a single hierarchy and that each object property can have only one value, consistent with the network model of RDF and its open world assumption. In another analogy with WebDAV (the DAV Searching and Locating (DASL) extension), Tupelo adds a query method supporting SPARQL. Authentication is managed via HTTPS and can be implemented using single sign-on mechanisms such as those developed at NCSA to leverage portal user databases or Shibboleth (Barton et al., 2006). Tupelo currently implements this via a JAAS Realm (Sun Developer Network, 2008).

**Client Libraries and User Interfaces**

Several client libraries have been developed for Tupelo, including low-level APIs in Java and Python. In Java, we have also created higher-level interfaces that provide direct support for manipulating metadata via Java Beans and their getter/setter methods, which we have then used in related projects to implement classes for common resources such as people, data files, documents, and workflows. Mechanisms to access Tupelo via JavaScript Object Notation (JSON) for web interfaces and via Adobe Air (Jobo.ZH, 2008) have also been created. Java Eclipse plug-in components that display raw and filtered versions of the network of relationships also exists and are being used, for example, to display provenance information (Figure 1). Tree and table plug-ins also exist and can likewise be configured to display desired subsets of the available metadata (such as the creator, creation date, and MIME type for a file-like display) and to browse specified relationships.

**Computational Inference and Indexing**

In addition to providing security and aggregation capabilities, Tupelo's Context mechanism acts as a general scoping mechanism. Decisions about aggregation and access control are in effect policies of the VO providing the Tupelo instance. Thus Contexts provide an opportunity to manage other types of policies and VO-level assumptions. Because Tupelo Contexts act as a broker for data and metadata operations, they can serve as a container for plug-in capabilities that augment data and metadata with inferred or computed information and/or perform specialized querying and indexing operations. This capability supports numerous e-Science use cases. For example, in the implementation of provenance capabilities, where it is straightforward to record direct ancestor relations, it is not possible to query for indirect ancestors, i.e. to find whether a given paper used a given set of observational inputs when there are intermediate derived data sets, with a SPARQL query over the metadata. Instead, we have implemented the transitive notion of ancestry within Tupelo by specifying simple rules described in a subset of the Semantic Web Rule Language (Horrocks et al., 2004) and executing these rules within a provenance Context that can be used to wrap underlying Contexts. During a query operation, this Context rewrites the query passed to underlying Contexts and performs the SWRL rules before returning the result. Thus, in a VO where the transitive nature of provenance is assumed, for example, by the OPM model, this assumption can be configured as part of the knowledge space infrastructure.

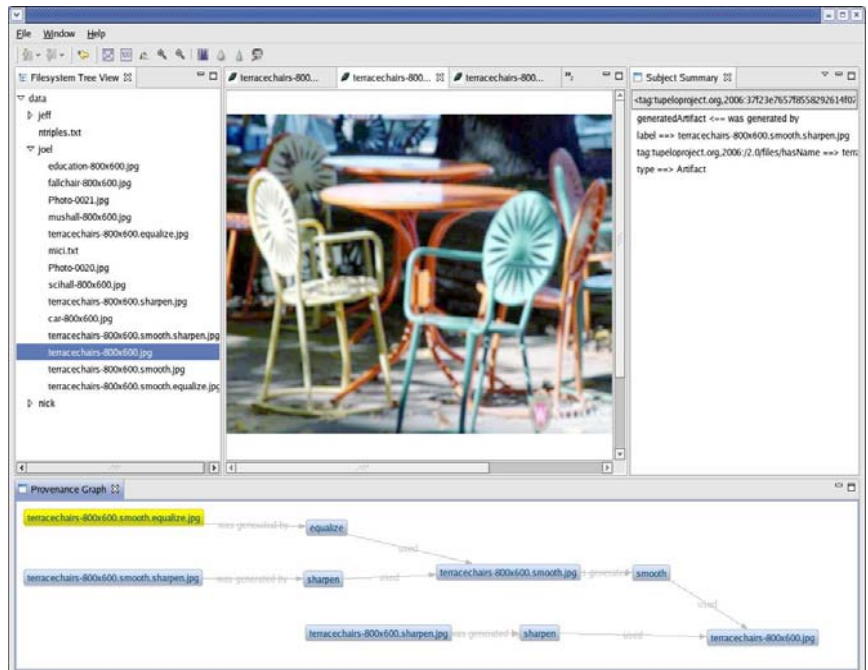

Figure 1. Java Eclipse components demonstrated in the context of an image viewer application: a tree view (left), retrieved content (center), metadata listing (right) and provenance graph (bottom). The CyberIntegrator environment (see text) provides similar data management and viewing capabilities within a larger framework to manage distributed computational workflow.

Tupelo Contexts may be used for purposes as simple as creating a full text-index, intercepting incoming write operations, and writing to an index before passing the write operation to a delegate, representing a fairly universal assumption (words have the same meaning across strings and documents). However, they can also be used to encode more complex notions, such as the spatial notion of "in" which can be derived mathematically from the location of one resource in terms of latitude, longitude, extent, and coordinate system relative to another resource. Tupelo allows one to ask queries that span different types of knowledge, e.g. to find people who have written papers about data from sensors within a given area. Tupelo can invoke the necessary inferences or computations using existing engines that efficiently implement the required data structures and algorithms. A wide range of knowledge could be encoded through this mechanism, ranging from fairly uncontroversial such as the geospatial and temporal relationships (sensor readings taken "during" a storm) to ones representing beliefs, policies, or shared assumptions of users, such as which books are "recommended", whether lossily compressed images should be considered equal/acceptable alternates to the original, or whether statistical correlations are acceptable (e.g. to identify "potential customer" relationships). In many ways, Tupelo's capability is comparable to the datablade concept (Olson, 1997), and similar tradeoffs between the utility of embedding the functionality within the data store and keeping it external will apply (Acker, Pieringer, & Bayer, 2005).

## Discussion

At NCSA we've undertaken a number of development and deployment efforts, using Tupelo as a repository behind other middleware and end-user applications, which show both the potential of a semantic content management abstraction and areas where additional development work or standardization will be needed. Collectively, these deployments serve hundreds of active researchers across a wide range of disciplines. One of the earliest tools to incorporate Tupelo is the Cyberintegrator (CI) workflow system (L. Marini, Minsker, Kooper, Myers, & Bajcsy, 2006). CI has more than 200 downloads and has been used in a wide range of research efforts ranging from academic analyses of urban run-off (Torres, 2007) and modeling of building earthquake fragilities (Elnashai & Lin, 2008), to industry-based geospatial data processing for risk analysis, cybersecurity log analysis, and aerodynamics modeling. CI's Tupelo-centric capabilities have been significant drivers in the selection of CI for these tasks. By using Tupelo to store all information about data, workflows, and its internal configuration, CI can support arbitrary data types and domain specific metadata. It can also record provenance that spans workflow sessions and group collaborations. Tupelo also enables CI to manage local and remote data through a common API and ignore the fact that data may actually be stored as files, at HTTP URLs, or, defined in terms of streams, derived dynamically from an underlying chunked storage representation of the stream (Rodriguez & Myers, 2008). CI also invokes Tupelo to manage annotations and tags about data, whole workflows, and tools (the individual components implementing the steps of the workflow). This information can be displayed as metadata and can also be used to reorganize data. Since the underlying repository does not have a single notion of hierarchy, CI can display data and tools arranged not only as a file-like tree in user defined directories, but organized by tag, mime-type, or relationship to a workflow.

More dramatically, since CI can directly store (or mirror) all data and tools required for a workflow to a remote repository, publication of the workflows to a remote server becomes a trivial matter of configuration and flagging workflows as available for remote execution. This latter capability forms the computational basis for our work on a Digital Synthesis Framework that can be used to dynamically create custom gateway-style web interfaces (Luigi Marini, Kooper, Myers, & Bajcsy, 2008). We have worked over the last year to implement a range of models in this system, working directly with researchers and educators, to produce interactive web environments including ones allowing exploration of streaming observations related to hypoxia (low-oxygen conditions) in Corpus Christi Bay in Texas, producing on-demand rainfall estimates (as "virtual rain gages") from streaming radar reflectivity measurements, and educational use of a cutting-edge plant growth model to understand agricultural yields under different farming practices and in the face of changing climate, Across these uses, the ability switch between desktop and web interfaces, between individual and group work, and manage data, metadata, and provenance coherently provides significant benefits in terms of ease of use. At the programming level, this flexibility has been critical to the rapid, incremental development of DSF capabilities themselves.

Tupelo has also been incorporated into our Liferay (liferay.com, 2008)-based Cybercollaboratory portal (Liu, McGrath, Myers, & Futrelle, 2007). NCSA has deployed more than a dozen of these portals serving more than 400 registered users. Usage information from collaborative portlets within the system are exposed via Tupelo for analysis (Rantannen, 2008) though we have not yet done so, be combined with information such as co-authorship, citation, and provenance information from other tools to analyze social networks. We are also developing a Tupelo-enabled document/data repository tool for this portal. As with CI, this choice allows the tool to show data residing on multiple servers (e.g. aggregating documents across sub-groups running separate portal infrastructures) and provides support for tagging and annotation. Further, because the data stored through CI and the portal are "just content" within a shared space, the document library can be used to explore data, provenance, and workflows created by CI, as well as their tags and annotations. Conversely, new annotations added through the portal become available within the CI's user interface.

The types of interoperability displayed in the uses above rely not just on Tupelo, but also to some extent on agreements about resource types and metadata terms between tools. For example, for multiple applications to share the notion of authorship, they must agree on the use of a common term for it, such as the Dublin Core (Dublin Core Metadata Initiative, 2006) term "creator". Toward this end we have adopted popular metadata sets including Dublin Core and Friend-of-a-Friend (FOAF) within tools we control. It should be noted that in cases where such agreement does not exist, metadata can still be displayed to users and Tupelo's inference capabilities can be used to map between islands of agreement. Further, developers can insulate themselves somewhat from vocabulary issues by using Tupelo's Bean API, which can translate Java objects into multiple RDF vocabularies.

Tupelo and the knowledge space abstraction do not necessarily solve the complex issues of sharing semantics. However, as with the use of XML and WebDAV, the use of RDF, the stack of semantic web languages, and the Tupelo protocol shifts the problem and simplifies the solution. Semantic agreements can be encoded declaratively as ontologies and transformation rules rather than being embedded in application code. Existing terms can be standardized without restricting the creation of new terms representing new logical models. As with the use of XSLT to transcode XML documents, one can define rules to map RDF vocabularies to meet VO conventions. Further, as is possible with the SAM WebDAV server, Tupelo allows such mappings to be performed on the server side on behalf of a VO rather than making such mappings the responsibility of individual tools. Given the range of potential deployment scenarios that currently require complex and error-prone negotiation over granularity, structure, serialization, and protocols such as the integration of information across institutional repositories being undertaken in the ECHO DEPository project (Rani et al., 2006), the existence of semantic tooling to automate agreements and a Context mechanism to implement agreements at the appropriate level in the infrastructure provide significant benefits over current practice.

Beyond the issues related to semantic agreement within and across VOs, which nominally apply equally to simple hierarchies of objects and more complex networks of information, there are also issues that apply more exclusively to the more complex case. Consider a scenario in which data might be generated by a mobile, off-network sensor and later transferred to a project repository and ultimately migrated from there to a long-term archive. In such a scenario, one would like to mint identifiers at the sensor source and then maintain them throughout the data lifecycle, to avoid costly negotiation across heterogeneous system boundaries (Oinn et al., 2006). As eloquently argued by Stephen Kunze, one would like identifiers to be persistent and work like URLs and provide guidance on where the data can be retrieved, leading to multipart, "actionable" URL identifiers such as the Archival Resource Key (ARK) (Kunze, 2003). ARKs encode the current data curator concatenated with a location-independent identifier. ARKs that share the same concatenated identifier can be interpreted as equivalent. We believe that the ARK model, harmonized with the Tag URI scheme (Kindberg & Hawke, 2005) to remove centralization of the minting process, as in Archival Resource Tags (Futrelle, 2006), provides a highly scalable and robust mechanism of citing e-Science data throughout its lifecycle. However, such two-part identifiers are not supported within standard semantic tools. We believe that managing ARK/ART-style identifiers can be done effectively using logic in a wrapper Context and will thus simplify data lifecycles in distributed e-Science.

Fine-grained access control in such a scenario raises additional issues. One can imagine policies that restrict access to data based on source, data type (e.g. photos), or provenance (data contributing to a conclusion). Supporting all policies together can lead to potential conflicts and undecidability. We believe that, in this case as well, Tupelo's support for wrapper Contexts provides an appropriate mechanism for defining and enforcing a given policy. For example, we are currently developing a simple policy supporting access control that is inherited along the directory hierarchy being implemented within our portal-based document library. In this case, the wrapper plug-in will enforce the constraint that directories are strict hierarchies and compute effective access control entries. This wrapper will be configurable to enforce security along any hierarchically-structured property.

## Conclusions

We have presented the concept of knowledge spaces that can represent both explicitly asserted data and metadata as well as inferred and computed consequences of those assertions. Further, we have reported on the development of the Tupelo middleware which implements a knowledge space abstraction using standard semantic web technologies and a protocol derived from current standards and modeled after the successful XML-oriented WebDAV protocol. A key feature of the Tupelo middleware and the knowledge space abstraction in general is the concept of aggregatable Contexts as a means to scope decisions ranging from simple configuration issues to complex and potentially problem-specific policies and assumptions of individual organizations. Our motivations in building Tupelo and the ways we see knowledge spaces benefitting e-Science are highlighted in the examples of real-world use of Tupelo given above and the exploration of Tupelo-related features of workflow and collaboration tools and underlying digital library and e-Science issues related to achieving semantic agreements and managing distributed, mobile, long-lived data. As noted, Tupelo draws heavily on existing technologies and our work has parallels with XML and relational approaches. However, we believe that it represents a relatively unique framework for e-Science developments that can simplify implementation of end-to-end provenance management and social networking over scientific data and workflows and can at least provide a framework in which to explore, both at scale and in operational contexts, the more complex issues of knowledge integration and the evolution of knowledge through scientific research.

## Acknowledgements

This material is based upon work supported by the National Science Foundation (NSF) under Award No. BES-0414259, BES-0533513, and SCI-0525308 and the Office of Naval Research (ONR) under award No. N00014-04-1-0437. Any opinions, findings, and conclusions or recommendations expressed in this publication are those of the author(s) and do not necessarily reflect the views of NSF and ONR.